\documentclass[10pt,twocolumn,letterpaper]{article}
\usepackage[T1]{fontenc}
\usepackage[pagenumbers]{cvpr}
\usepackage{array,multirow}
\usepackage{dsfont,balance}
\definecolor{cvprblue}{rgb}{0.21,0.49,0.74}
\usepackage[breaklinks,colorlinks,allcolors=cvprblue]{hyperref}

\title{TRAIL: Expert-Aligned and Terrain-Robust\\
Imitation Learning for Humanoid Robots}
\author{%
  Lurui Wang\textsuperscript{a},
  Yilin Chen\textsuperscript{b},
  Pengjie Xu\textsuperscript{c},
  Zhaojun Deng\textsuperscript{d}\thanks{Corresponding author. E-mail: \href{mailto:xupj1992@163.com}{xupj1992@163.com}}\\[0.55em]
  {\small\textsuperscript{a}Department of Robotics, Johns Hopkins University, Baltimore, MD 21218, USA}\\
  {\small\textsuperscript{b}Hubei Key Laboratory of Intelligent Robot (Wuhan Institute of Technology)),}\\
  {\small Wuhan City, Hubei Province, 430073, China}\\
  {\small\textsuperscript{c}School of Mechanical Engineering, Shanghai Jiao Tong University, Shanghai, 200240, China}\\
  {\small\textsuperscript{d}College of Surveying and Geo-Informatics, Tongji University, Shanghai, 200092, China}\\[0.26em]
  {\small\textsuperscript{a}E-mail: \href{mailto:lwang255@jh.edu}{lwang255@jh.edu}\qquad
    \textsuperscript{d}E-mail: \href{mailto:xupj1992@163.com}{xupj1992@163.com}}%
}
\hypersetup{pdftitle={TRAIL: Expert-Aligned and Terrain-Robust Imitation Learning for Humanoid Robots},
  pdfauthor={Lurui Wang, Yilin Chen, Pengjie Xu, Zhaojun Deng}}

\begin{document}
\maketitle

\begin{abstract}
Imitation learning efficiently reproduces expert behavior. However, embodiment-mismatched demonstrations create a gap between human data and a half-body robot. Human full-body trajectories do not transfer to the robot directly. Many recent methods hard-learn from such mismatched data. They force-fit policies and accept brittle performance. This work takes a different route. This paper proposes an end-to-end framework that bridges embodiment mismatch in imitation learning. The projection stage aligns cross-embodiment kinematics and structure and generates pseudo-expert trajectories. The training stage performs adversarial imitation learning and policy optimization on these trajectories. The procedure integrates resampling to refine data quality. It also applies terrain-parameter randomization to improve robustness and generalization. The method is evaluated in MuJoCo under multi-terrain settings. Results show stable walking, consistent foot placement, and a low fall rate.
\end{abstract}

{\small\raggedright\noindent\textbf{Keywords:} Multi-terrain locomotion; Embodiment mismatch; Reinforcement learning; Cross-embodiment projection.\par}

\begin{figure*}[t]
\centering
\includegraphics[width=\linewidth]{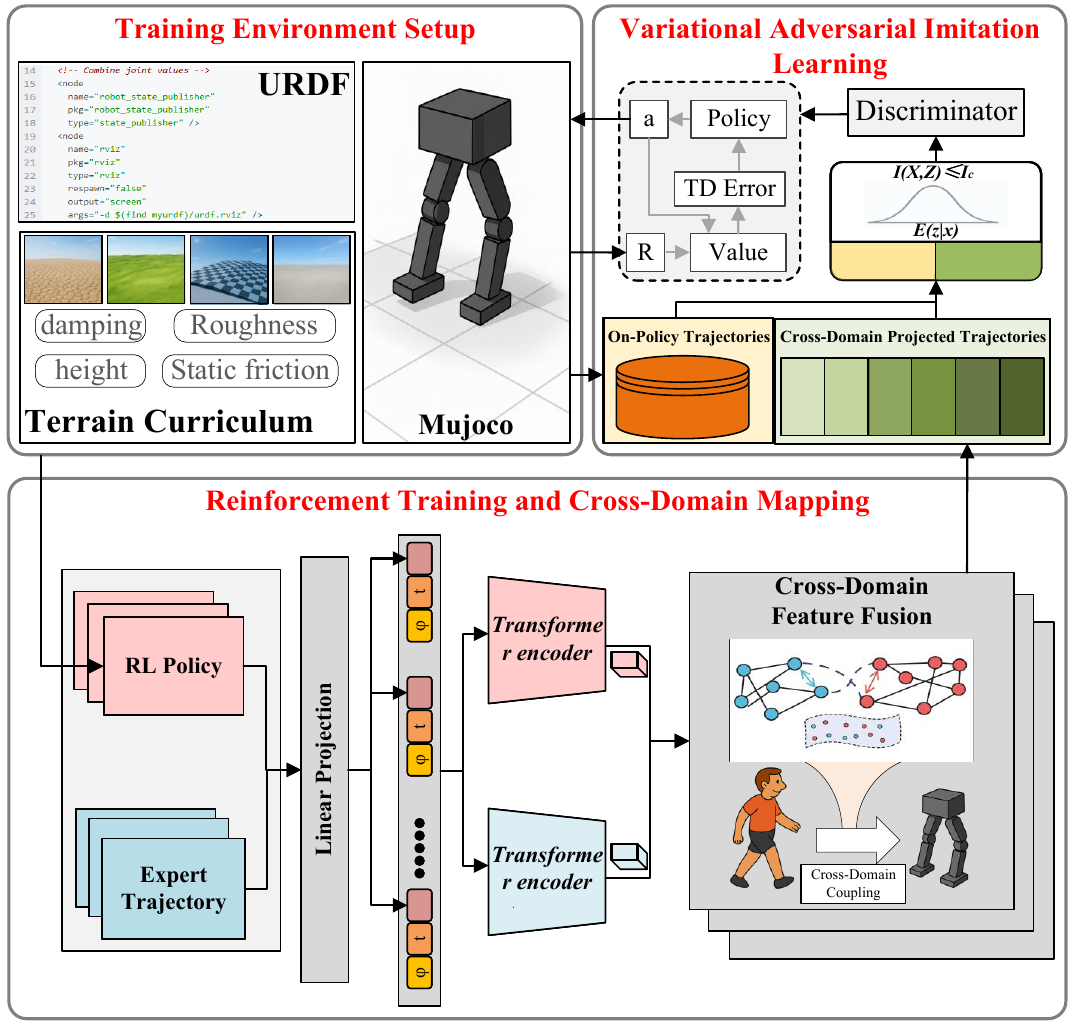}
\caption{Overview of the Projection--Training framework.
(a) Reinforcement training and cross-domain mapping (Module~I): learns human--robot trajectory alignment via FGW projection with dual Transformer encoders.
(b) Training environment setup and VAIL imitation learning (Module~II): VAIL-based policy training with terrain curriculum, resampling, and TRPO refinement.}
\label{fig:framework}
\end{figure*}

\section{Introduction}
Classical model-based controllers provide theoretical stability for legged robots~\cite{1,2,3} but rely on precise dynamics models and struggle in high-DOF, nonlinear, and impact-rich settings. Deep reinforcement learning (RL) has emerged as an alternative, training policies on full-order dynamics and often outperforming classical methods in multi-contact environments. Hwangbo et al. trained a policy for the ANYmal robot that maintained balance under large disturbances and achieved robust, fast motion~\cite{4,5}. However, RL depends on hand-crafted reward functions that fail to capture motion style and task intent. Imitation learning (IL) mitigates this limitation by learning directly from expert demonstrations; for example, Peng et al.~\cite{6} reproduced agile animal motions for Laikago, enabling stable walking and turning at medium speed.
However, imitation learning is limited by the consistency of its demonstration trajectories, since embodiment mismatch prevents direct transfer from demonstrations to the student robot.
In this paper, cross-embodiment refers to the mismatch between the expert and the learner in three coupled aspects: morphology, control interface, and state--action space. Morphology mismatch includes differences in kinematic topology, link proportions, and available degrees of freedom; control-interface mismatch includes differences in actuation layout, command semantics, and feasible action ranges; and state--action mismatch arises because the expert and the robot do not share a common observation or control representation. Consequently, the challenge is not merely to retarget kinematics frame by frame, but to construct supervision that remains executable under the target robot's dynamics, contact patterns, and embodiment constraints. As a result, policies degrade under out-of-distribution conditions~\cite{7,8,9,10,11,12}.

Existing work follows two main routes to mitigate this issue. The first route addresses cross-domain transfer by performing alignment inside adversarial imitation learning. Representative methods include Domain Adaptive Imitation Learning~\cite{13}, which learns explicit state--action correspondences via Generative Adversarial MDP Alignment and then composes them for zero-shot transfer within an adversarial align--adapt pipeline, and Out-of-Dynamics Imitation Learning (OOD-IL)~\cite{10}, which first clusters demonstrations to handle multimodality and then uses Generative Adversarial Imitation Learning (GAIL) to estimate transferability and re-weight demonstrations during policy learning. In contrast, our method performs trajectory alignment as a standalone, pre-training step---before any adversarial imitation updates---so no in-loop discriminator-driven alignment is required. Accordingly, the contribution of TRAIL is not a loose combination of existing modules. Its core novelty lies in a projection-before-training paradigm for cross-embodiment imitation. Rather than estimating correspondence inside adversarial policy learning, TRAIL first learns task-relevant temporal embeddings that capture transition structure shared by human and robot motion, then solves FGW in this shared embedding space instead of the raw state space, and finally converts the resulting soft correspondence into executable robot supervision through barycentric projection over the robot trajectory library. This design explicitly decouples correspondence estimation from policy optimization, turns embodiment alignment into a standalone pre-training problem, and allows downstream imitation learning to operate on executable pseudo-expert trajectories rather than directly on mismatched human demonstrations.

The second route treats embodiment mismatch as noise in demonstrations. It trains an imitation policy that is insensitive to this noise. Noise Injection for Robust Imitation Learning~\cite{14} injects optimized noise into the expert controller during data collection to elicit recovery actions and reduce covariate shift. It keeps training off-policy while matching the states the learned policy will visit. It improves safety and computation during training. Robust Inverse Reinforcement Learning~\cite{15} models the dynamics gap as an uncertainty set over transition kernels and solves a min--max Maximum Causal Entropy Inverse Reinforcement Learning objective. It bounds performance loss by the $L_1$ distance between expert and learner dynamics. It yields policies that remain stable as the mismatch grows and provides a knob to tune robustness. In other words, these methods do not align domains but make imitation learning more tolerant to imperfect or mismatched demonstrations.
In summary, most prior work designs new imitation-learning mechanisms. Those mechanisms mainly distill usable signal from such demonstrations. This paper takes a data-side route. It explicitly synthesizes executable substitute trajectories under the target embodiment and environment. It expands the effective domain of demonstrations without changing the core imitation-learning paradigm. Action-Constrained Imitation Learning (ACIL)~\cite{16} addresses imitation when the imitator faces action constraints and the expert does not. ACIL eliminates occupancy-measure distortion through Dynamic Time Warping--guided trajectory alignment using Model Predictive Control, generating surrogate demonstrations for downstream imitation learning. Building on this idea, this paper extends trajectory operations to handle embodiment differences and terrain variation, generating substitute trajectories that match target conditions.
This paper presents a cross-embodiment trajectory generation framework that generalizes demonstrations to new dynamics and environments while preserving the imitation-learning paradigm. Through resampling and terrain randomization, it achieves robust and stable performance under unseen conditions. The main contributions are as follows.
\begin{itemize}
\item An end-to-end cross-embodiment trajectory pipeline.
It employs a two-tower encoder with a shared head to extract human--robot temporal representations.
It performs Fused Gromov--Wasserstein (FGW) alignment to obtain the projection matrix,
mapping human trajectories to robot-executable pseudo-expert sequences under dynamics and environment mismatch.
\item A robust imitation-learning procedure.
It trains the policy with Variational Adversarial Imitation Learning (VAIL),
integrates resampling and curriculum-based terrain randomization to enhance robustness,
and validates the method on multi-terrain simulation.
\end{itemize}
\section{Methodology}
\subsection{Overview}
The proposed TRAIL framework embodies an end-to-end ``Projection--Training'' scheme that bridges expert demonstrations and terrain-robust humanoid control. It consists of two modules: an end-to-end cross-embodiment trajectory pipeline and a robust imitation-learning procedure, as illustrated in Fig.~\ref{fig:framework}.
Module I comprises four components. A reinforcement-learning policy is first trained across multiple environments to collect diverse, contact-robust reference trajectories. Temporal information is then encoded with Sinusoidal positional embeddings. A two-tower Transformer encoder with a shared head, jointly trained with a transition-discrimination task, extracts unified temporal features. Finally, the projection matrix is obtained via FGW, mapping human trajectories to robot-executable pseudo-expert sequences.
Module II consists of four stages. The policy is first trained with VAIL, followed by terrain randomization with a curriculum to increase complexity. High-quality trajectories are then resampled to expand capability, and the policy is finally refined using Trust Region Policy Optimization (TRPO).
\subsection{End-to-end cross-embodiment trajectory pipeline}
\subsubsection{multi-terrain trajectory collection via reinforcement learning}
\label{sec:rl_collection}
Separate reinforcement-learning policies are trained in simulation across diverse terrains using TRPO to collect dynamically feasible walking trajectories. The observation space includes the robot state $o_t$~\eqref{eq:obs} and a gait clock signal designed following Singh et al.~\cite{17}.
\begin{equation}
o_t = (\phi, \omega, q, \dot{q}),
\label{eq:obs}
\end{equation}
where $\phi, \omega$ denote the base orientation and angular velocity,
and $q, \dot{q}$ denote the joint positions and velocities.
\begin{equation}
clock = \left\{
\sin\!\left(\frac{2\pi \phi}{L}\right),\,
\cos\!\left(\frac{2\pi \phi}{L}\right)
\right\},
\label{eq:clock}
\end{equation}
The policy output is denoted by $a_t$~\eqref{eq:action}.
The term $a\,\delta\phi$ handles irregular terrains where strict periodicity is hard to maintain.
Its design follows Singh et al.~\eqref{eq:phase_update}.

\begin{equation}
a_t = \left(q^{des},\, \dot{q}^{des},\, \tau^{res},\, a_{\delta \phi}\right),
\label{eq:action}
\end{equation}
$q^{\mathrm{des}}$ denotes the desired joint positions.
$\dot{q}^{\mathrm{des}}$ denotes the desired joint velocities.
$\tau^{\mathrm{res}}$ denotes the residual torques.
$a_{\delta\phi}$ denotes the policy-predicted phase offset used to adjust the gait cycle.
\begin{equation}
\phi_{t+1} = \phi_t + \mathrm{clip}(a_{\delta \phi}, -\Delta, +\Delta) + 1,
\label{eq:phase_update}
\end{equation}
$\phi_t$ denotes the phase value at time $t$.
$a_{\delta \phi}$ denotes the policy-predicted phase-offset action.
$\Delta$ denotes the lower and upper bounds of the offset.
The reward design follows prior humanoid locomotion frameworks~\cite{18,19} but is adapted for half-body forward walking by reweighting terms, removing upper-body components, and keeping forward velocity as the sole objective. Two regularizers are added: an action-smoothing term (Eq.~\eqref{eq:r_act}) to reduce jitter and a torque penalty (Eq.~\eqref{eq:r_tau}) to limit actuation.
\begin{equation}
r_{\mathrm{act}} =
\exp\!\left(
-\alpha \cdot
\frac{1}{n}
\sum_{i=1}^{n}
\left| a_i - a_i^{\mathrm{prev}} \right|
\right),
\label{eq:r_act}
\end{equation}
$a_i$ and $a_i^{\text{prev}}$ denote the actions of joint $i$ at the current and previous time steps, respectively. The parameter $\alpha$ controls the sensitivity to action changes.
\begin{equation}
r_{\tau} =
\exp\!\left(
-\beta \cdot
\frac{1}{n}
\sum_{i=1}^{n}
\left| \tau_i - \tau_i^{\mathrm{prev}} \right|
\right),
\label{eq:r_tau}
\end{equation}
$\tau_i$ and $\tau_i^{\mathrm{prev}}$ denote the torque of joint $i$ at the current and previous time steps.%
$\beta$ serves as a scaling factor that controls the penalty strength.%
\subsubsection{Transformer-based Cross-Domain Semantic Alignment}
Inspired by shared-weight architectures in multi-modal and cross-domain learning~\cite{20,21,22,23}, this module shares the encoder, following prior work showing that shared projections align features from different sources into a unified semantic space.
As shown in Fig.~\ref{fig:framework}, the training architecture uses a two-tower encoder with a shared MLP head, where the human and robot encoders extract features from their respective trajectories.%
Then, the shared projection head $g(\cdot;\theta)$ maps both feature streams to embeddings $Z$, with shared parameters $\theta$ placing the two domains in the same semantic space.%
As a result, the aligned embeddings provide a direct basis for Fused Gromov--Wasserstein (FGW) to compute the coupling matrix.
The training criterion we use is a task-relevant embedding objective inspired by Franzmeyer \textit{et~al.}~\cite{24}.
Specifically, we define a triplet of random variables $(S, S', Y)$ from expert demonstrations.
We construct real pairs $\mathcal{T}_{\mathrm{real}} = \{(s_t, s_{t+1})\}$ and random pairs $\mathcal{T}_{\mathrm{random}} = \{(s_i, s_j)\}$ by sampling $i$ and $j$ independently from the same set.%
We set the binary label $Y$ to indicate pair type, with $Y = 1$ for real and $Y = 0$ for random.%
For any time-invariant embedding $f: S \to Z$, we obtain $(Z, Z') = (f(S), f(S'))$ and thus induce $(Z, Z', Y)$.%
We train the encoder to maximize the mutual information~\eqref{eq:mi_obj} between $Z$ and $Z'$ for positive pairs.%
This objective learns task-relevant embeddings by distinguishing real from random transitions.%
\begin{equation}
\begin{aligned}
&\arg\max_{f}\, I\!\left((Z, Z');\, Y\right) \\
&\quad= \arg\max_{f}\, I\!\left((f(S), f(S'));\, Y\right),
\end{aligned}
\label{eq:mi_obj}
\end{equation}
In implementation, we adopt a transition-discrimination training scheme. In practice, this discriminative objective is equivalent to maximizing a common lower bound on $I((Z, Z'); Y)$~\cite{25}. Specifically, we treat consecutive pairs $(z_t, z_{t+1})$ as positive samples, and we treat non-adjacent pairs $(z_t, z_j)$ with $(j \neq t+1)$ as negative samples. Then, each pair is passed through a binary MLP head with a Sigmoid output predicting real (1) or random (0) transitions, and the predictor is trained using the binary cross-entropy loss in Eq.~\eqref{eq:bce1}.
\begin{equation}
\begin{aligned}
\mathcal{L}_{\mathrm{BCE}} &= \mathbb{E}_{(s_t, s_{t+1}) \sim \mathcal{D}} \Big[
-y \log \sigma\!\big(h(s_t, s_{t+1})\big) \\
&\quad -(1-y)\log\!\big(1-\sigma(h(s_t,s_{t+1}))\big)\Big].
\end{aligned}
\label{eq:bce1}
\end{equation}
$\sigma(\cdot)$ denotes the Sigmoid function, and $\mathbb{E}[\cdot]$ denotes the expectation over the corresponding sample set.
The shared MLP head discriminates transitions from both robot and human domains, with the supervised loss defined as the sum of two binary cross-entropy terms Eq.~\eqref{eq:shared_loss}.
\begin{equation}
\mathcal{L}_{\mathrm{shared}}
=
\mathcal{L}_{\mathrm{BCE}}^{\mathrm{human}}
+
\mathcal{L}_{\mathrm{BCE}}^{\mathrm{robot}}
\label{eq:shared_loss}
\end{equation}
Specifically, $L^{(h)}_{\mathrm{CE}}$ and $L^{(r)}_{\mathrm{CE}}$ denote the binary cross-entropy losses computed in the human domain and the robot domain, respectively.
\subsubsection{Cross-domain projection and pseudo-expert generation with Fused Gromov--Wasserstein (FGW)}
After obtaining a unified embedding, the pseudo-expert trajectory is projected into the robot action space via correspondence estimation and barycentric projection. Given human demonstrations $\tau_h = \{x_j\}_{j=1}^m$ and the robot action library $\tau_r = \{y_i\}_{i=1}^n$, the shared-head two-tower encoder provides embeddings $f_h(\cdot)$ and $f_r(\cdot)$. Correspondence is established by solving the FGW objective, which balances attribute discrepancy and structure preservation; see Eq.~\eqref{eq:ot1}~\cite{26}.
\begin{equation}
\begin{aligned}
\hat{T} &= \arg\min_{T \in \mathcal{U}(\mathbf{w}_{\tau_r}, \mathbf{w}_{\tau_h})}
\Bigl[\underbrace{\lambda_{\mathrm{attr}}\sum_{i=1}^{n}\sum_{j=1}^{m} M_{ij}T_{ij}}_{\text{attribute discrepancy term}} \\
&{}+\underbrace{\lambda_{\mathrm{str}}
\sum_{i,k=1}^{n}\sum_{j,l=1}^{m}
\bigl(C_{\tau_r}(i,k)-C_{\tau_h}(j,l)\bigr)^2 T_{ij}T_{kl}}_{\text{structure-preservation term}}\Bigr].
\end{aligned}
\label{eq:ot1}
\end{equation}
$T \in \mathcal{U}(\mathbf{w}_{\tau_r}, \mathbf{w}_{\tau_h})$ denotes the \textit{coupling matrix}, and its marginals match the frame weights $\mathbf{w}_{\tau_r}$ and $\mathbf{w}_{\tau_h}$ in two domains. $M_{ij}$ denotes the \textit{attribute term} that measures the discrepancy between expert frame $x_i$ and $y_j$ in the shared feature space, and it is defined as $M_{ij} = \| f_r(y_i) - f_h(x_j) \|_2^2$. $\mathbf{C}_{\tau_r}$ and $\mathbf{C}_{\tau_h}$ denote the \textit{intra-domain distance matrices}; specifically, $\mathbf{C}_{\tau_r}(i, k) = \| y_i - y_k \|_2^2$, $\mathbf{C}_{\tau_h}(j, l) = \| x_j - x_l \|_2^2$.
The optimal solution to this optimization problem is the \textit{coupling matrix} $\hat{T}$. Its $(i,j)$-th entry represents the \textit{soft correspondence weight} between the $j$-th frame of $\tau_r$ and the $i$-th frame of $\tau_r$.
After obtaining the coupling matrix $\hat{T}$, we apply \textit{barycentric projection} to map expert frames into the robot action space by row-normalizing $\hat{T}$ to get weights $\hat{T}_{ij}$~\eqref{eq:coupling_weights}.
\begin{equation}
\tilde{T}_{i:} =
\frac{T_{i:}}{\sum_{j}T_{ij}}
\label{eq:coupling_weights}
\end{equation}
and it then represents each pseudo-expert action as a convex combination of the robot action atoms $\tau_r = \{ y_i \}_{i=1}^n$~\eqref{eq:projected_trajectory}.
\begin{equation}
\begin{aligned}
\hat{u}_{i} &= \sum_{j=1}^{m} \hat{T}_{ij}\, y_{j}, \\
\tau_{\mathrm{proj}} &= (\hat{u}_{1}, \dots, \hat{u}_{n})
\end{aligned}
\label{eq:projected_trajectory}
\end{equation}
This step produces an executable sequence in the robot action space.
\subsection{Robust imitation-learning procedure}
\label{sec:imitation}
\subsubsection{Training Setup}
Policies are trained across multiple terrains $\{\mathcal{E}_i\}_{i=1}^{M}$.
For each terrain, the pipeline constructs expert and generated buffers from
$\tau_{\mathrm{proj}}$ and sampled trajectories, sets up a discriminator
$D_i$ and encoder $E_i$, performs adversarial training with the VAIL objective~\cite{27},
and updates the policy $\pi_i$ using TRPO.
\subsubsection{Buffer construction (Expert / Generated Buffers)}
Each terrain $i$ constructs two buffers: the \textit{expert buffer} $B_{\mathrm{exp}}^{(i)}$ and the \textit{generated buffer} $B_{\mathrm{gen}}^{(i)}$.
Here, $B_{\mathrm{gen}}^{(i)}$ corresponds to \textbf{on-policy trajectories} collected from the current policy,
while $B_{\mathrm{exp}}^{(i)}$ corresponds to \textbf{cross-domain projected trajectories} serving as expert data.
Specifically, $B_{\mathrm{exp}}^{(i)}$ contains the pseudo-expert trajectories $\tau_{\mathrm{proj}}$ produced in Cross domain projection and thus serves as the target distribution,
whereas $B_{\mathrm{gen}}^{(i)}$ contains trajectories collected from the interaction between the current policy $\pi_i$ and environment $\mathcal{E}_i$.
Mini-batches from both buffers are sampled at a fixed ratio and fed to the discriminator for imitation rewards,
while top-performing segments are added to $B_{\mathrm{exp}}^{(i)}$ to refine the target distribution.
\subsubsection{Variational Adversarial Imitation Learning (VAIL)}
Each terrain $i$ adopts the VAIL framework~\cite{27} for adversarial training. Samples from $B_{\mathrm{exp}}^{(i)}$ and $B_{\mathrm{gen}}^{(i)}$ are first encoded into latent variables $z$ by $E_i(z \mid x)$, and the discriminator $D_i$ then classifies them in the latent space. Meanwhile, a mutual-information budget on $I(x, z)$ constrains the discriminator and regularizes training. The joint objective for terrain $i$ over the discriminator--encoder pair is given in Eq.~\eqref{eq:vail_objective}.
\begin{equation}
\begin{aligned}
&\min\max_{D_i,\,E_i,\,\beta_i \ge 0}\;
  \mathbb{E}_{x \sim B_{\mathrm{exp}}^{(i)}} \\
&\qquad \mathbb{E}_{z \sim E_i(z\mid x)}\big[-\log D_i(z)\big] \\
&\quad +\mathbb{E}_{x \sim B_{\mathrm{gen}}^{(i)}}
  \mathbb{E}_{z \sim E_i(z\mid x)}\big[-\log\big(1 - D_i(z)\big)\big] \\
&\quad +\beta_i\left(
  \mathbb{E}_{x \sim \tilde{B}^{(i)}}
  \mathrm{KL}\!\left[E_i(z\mid x)\,\|\,r(z)\right]
  - I_c
  \right)
\end{aligned}
\label{eq:vail_objective}
\end{equation}
$\tilde{B}^{(i)}$ denotes the equal mixture of the expert and generated buffers; defined as
$\tilde{B}^{(i)}=\frac{1}{2}B_{\mathrm{exp}}^{(i)}+\frac{1}{2}B_{\mathrm{gen}}^{(i)}$.
The prior $r(z)=\mathcal{N}(0,I)$ denotes a standard normal distribution with zero mean and identity covariance.
The mutual-information budget $I_c$ in the constraint limits information transmitted by the encoder.
The multiplier $\beta_i$ is updated by \textbf{dual gradient descent} to keep the constraint active, and its update rule is given in Eq.~\eqref{eq:dual_update}.
\begin{equation}
\begin{aligned}
\beta_i &\leftarrow \max\!\Bigl(0,\;\beta_i + \alpha_\beta\Bigl[ \\
&\quad\mathbb{E}_{x \sim \tilde{B}^{(i)}}
\mathrm{KL}\!\left[E_i(z|x)\,\|\,r(z)\right]
- I_c\Bigr]\Bigr)
\end{aligned}
\label{eq:dual_update}
\end{equation}
$\alpha_{\beta}$ denotes the step-size coefficient in dual gradient descent, and it controls the magnitude of the update.
The instantaneous reward defined by the discriminator is given by Eq.~\eqref{eq:imitation_reward}.
\begin{equation}
r_i(x)
= -\log\!\left(1 - D_i\!\big(\mu_{E_i}(x)\big)\right)
\label{eq:imitation_reward}
\end{equation}
$\mu_{E_i}(x)$ denotes the \textbf{mean mapping} of the encoder.
Then, the pipeline \textbf{performs policy optimization} with TRPO for terrain $\mathcal{E}_i$ and \textbf{obtains} the policy $\pi_i$.
Finally, the procedure \textbf{runs in parallel} for $i=1,\ldots,M$ and \textbf{yields} the policy set $\{\pi_i\}_{i=1}^M$.
\subsubsection{Resampling and Adaptive Curriculum Learning}
To further enhance robustness, terrain parameters are randomly perturbed, and task difficulty is adjusted online via Adaptive Curriculum Learning (ACL). The survival step count $l_t$ is given in Eq.~\eqref{eq:survival_step}.
\begin{equation}
\begin{aligned}
\ell_t &= \mathbf{1}\!\left\{
  \begin{aligned}
  &\text{episode not terminated}\\
  &{}\wedge\;v_{\mathrm{pr}}(s_{t+1}) > 0
  \end{aligned}
  \right\},\\
&\qquad t = 0, \ldots, H - 1.
\end{aligned}
\label{eq:survival_step}
\end{equation}
$\mathds{1}\{\cdot\}$ denotes the indicator function.
$v_{\mathrm{pr}}(s_{t+1})$ denotes the forward-progress velocity evaluated at state $s_{t+1}$.
Accordingly, we define two criteria given in Eq.~\eqref{eq:success_criteria}.
\begin{equation}
\begin{aligned}
\mathrm{Succ}_H(\tau) &= \mathbf{1}\!\left\{\frac{1}{H}\sum_{t=0}^{H-1} \ell_t \ge \alpha\right\},
\quad \alpha = 0.2,\\[4pt]
\mathrm{Full}_H(\tau) &= \mathbf{1}\!\left\{\frac{1}{H}\sum_{t=0}^{H-1} \ell_t = 1\right\}.
\end{aligned}
\label{eq:success_criteria}
\end{equation}
Two criteria are used.
If three consecutive rollouts succeed ($\mathrm{Full}_H(\tau)=1$), the scheduler increases terrain difficulty;
if three fail ($\mathrm{Succ}_H(\tau)\neq1$), it decreases it.
Once stable success is reached, a small set of successful trajectories is added to the expert buffer for self-imitation.
\subsection{Trust Region Policy Optimization (TRPO)}
The policy is updated using \textbf{Trust Region Policy Optimization (TRPO)}~\cite{28},
where the previous policy $\pi_{\mathrm{old}}(a \mid s)$ samples trajectories
and estimates the advantage $A_{\mathrm{old}}(s,a)$ in Eq.~\eqref{eq:advantage}.
\begin{equation}
\begin{aligned}
\hat{A}_t^{\mathrm{old}}
&= \sum_{l=0}^{T-t-1}\gamma^{l}\,
\delta_{t+l}^{\mathrm{old}}, \\
\delta_t^{\mathrm{old}}
&= r_t + \gamma V^{\mathrm{old}}(s_{t+1})
- V^{\mathrm{old}}(s_t)
\end{aligned}
\label{eq:advantage}
\end{equation}
$t$ denotes the timestep.
$T$ denotes the length of the trajectory segment.
$s_t$ denotes the observed state.
$V(\cdot)$ denotes the value estimator.
$\gamma \in (0,1)$ denotes the discount factor.
The surrogate objective in Eq.~\eqref{eq:surrogate} is then maximized.
\begin{equation}
\begin{aligned}
L_{\mathrm{old}}(\theta)
&= \mathbb{E}\!\left[r_{\theta}(s,a)\,A_{\mathrm{old}}(s,a)\right], \\
r_{\theta}
&= \frac{\pi_{\theta}(a|s)}{\pi_{\mathrm{old}}(a|s)}
\end{aligned}
\label{eq:surrogate}
\end{equation}
$\pi_{\theta}(a|s)$ denotes the new policy.
$r_{\theta}$ denotes the preference ratio.
$\mathbb{E}[\cdot]$ denotes an average over samples.
However, to avoid overly large updates, an average KL constraint has been imposed, given in Eq.~\eqref{eq:kl_constraint}.
\begin{equation}
\begin{aligned}
&\max_{\theta}\; L_{\mathrm{old}}(\theta) \\
&\text{s.t.}\quad
\mathbb{E}\!\left[
D_{\mathrm{KL}}\!\left(
\pi_{\mathrm{old}}(\cdot|s)
\Vert
\pi_{\theta}(\cdot|s)
\right)
\right]
\leq \delta
\end{aligned}
\label{eq:kl_constraint}
\end{equation}
$D_{KL}$ measures the distance between the old and new policies. $\delta$ sets the trust-region radius. Assuming the step size $\beta$ is small enough to satisfy the quadratic KL approximation give in Eq.~\eqref{eq:kl_approximation}.
\begin{equation}
\mathbb{E}[D_{\mathrm{KL}}(\pi_{\mathrm{old}}(\cdot|s)\parallel\pi_{\theta}(\cdot|s))]\approx\frac{1}{2}\beta^{2}n^{\mathsf{T}}Fn
\label{eq:kl_approximation}
\end{equation}
$\mathbf{n}$ denotes the natural-gradient direction.
$\mathbf{F}$ denotes the \textbf{Fisher information matrix} defined by
$\mathbf{F}=\mathbb{E}[\nabla\log\pi_{\theta}\,\nabla\log\pi_{\theta}^{\top}]$.
Therefore, to satisfy the constraint, the step size obeys Eq.~\eqref{eq:step_size}, and the parameters update as Eq.~\eqref{eq:policy_update}.
\begin{equation}
\beta\leq\sqrt{\frac{2\delta}{n^{\mathsf{T}}Fn}}
\label{eq:step_size}
\end{equation}
\begin{equation}
\theta_{\mathrm{new}}=\theta_{\mathrm{old}}+\beta n
\label{eq:policy_update}
\end{equation}
\section{Simulation}

\begin{figure*}[t]
\centering
\includegraphics[width=\linewidth]{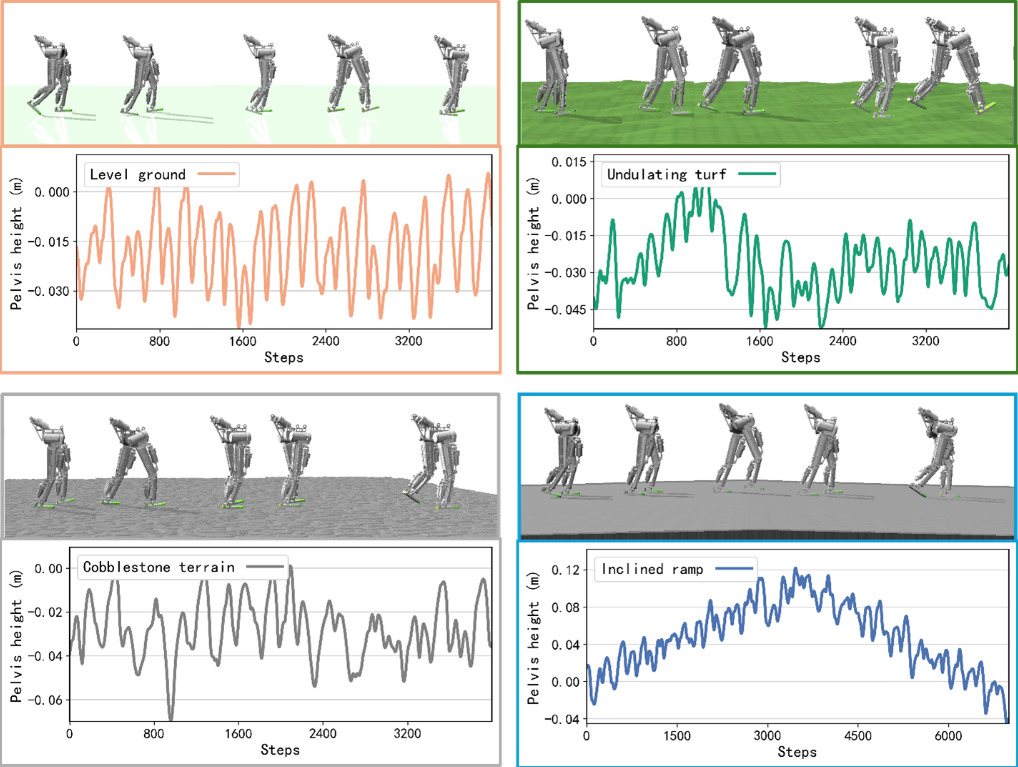}
\caption{Robot locomotion across different terrains.}
\label{fig:locomotion}
\end{figure*}
This section investigates two questions: (i) whether projection-based expansion of expert demonstrations can be aligned effectively to a half-body humanoid, and (ii) whether the resulting policy remains robust across diverse terrains. We adopt a unified training--evaluation protocol and conduct a series of controlled experiments, detailed below.
\begin{table}[tbp]
\centering
\small
\caption{Reward Function.}
\label{tab:reward}
\setlength{\tabcolsep}{4pt}\renewcommand{\arraystretch}{1.08}
\begin{tabular}{@{}>{\raggedright\arraybackslash}p{.36\linewidth}>{\raggedright\arraybackslash}p{.38\linewidth}r@{}}
\toprule
\textbf{Category} & \textbf{Term} & \textbf{Weight} \\
\midrule
Bipedal walking (clock-based)
& Foot force  & 0.26 \\
& Foot speed  & 0.26 \\[2pt]
Objective
& Forward velocity & 0.22 \\[2pt]
Safety and Realism
& Root height & 0.06 \\
& Nominal posture & 0.06 \\
& Joint velocities & 0.06 \\
& Action smoothness & 0.04 \\
& Torque smoothness & 0.04 \\
\bottomrule
\end{tabular}
\end{table}
\subsection{Experiments setup}
This study trains a half-body humanoid robot in the MuJoCo~2.3.6 simulator on four terrains---Level~Ground, Undulating~Turf, Cobblestone, and Inclined~Ramp---using a residual-plus-PD controller to achieve stable and adaptive locomotion.
For reproducibility, this subsection consolidates experimental settings from the \textit{Methodology} section:
the reinforcement-learning (RL) process corresponds to Section~\ref{sec:rl_collection}, \textit{Multi-Terrain Trajectory Collection via Reinforcement Learning}, which generates expert reference motions under diverse terrain conditions;
the imitation-learning (IL) process corresponds to Section~\ref{sec:imitation}, \textit{Robust Imitation-Learning Procedure}, which transfers these expert behaviors into a student policy for deployment.
Table~\ref{tab:reward} lists reward categories and weights (RL), Table~\ref{tab:policy_inputs} summarizes policy inputs (RL and IL), and Table~\ref{tab:policy_outputs} reports policy outputs.

\begin{table}[tbp]
\centering
\small
\caption{Policy inputs.}
\label{tab:policy_inputs}
\setlength{\tabcolsep}{4pt}\renewcommand{\arraystretch}{1.08}
\begin{tabular}{@{}>{\raggedright\arraybackslash}p{.26\linewidth}>{\raggedright\arraybackslash}p{.26\linewidth}ccc@{}}
\toprule
\textbf{Category} & \textbf{Observation} & \textbf{Dimension} & \textbf{RL} & \textbf{IL} \\
\midrule
\multirow{4}{*}{Robot state}
& Root positions & 6 & $\checkmark$ & $\checkmark$ \\
& Root velocities & 6 & $\checkmark$ & $\checkmark$ \\
& Joint positions & 10 & $\checkmark$ & $\checkmark$ \\
& Joint velocities & 10 & $\checkmark$ & $\checkmark$ \\[2pt]
External state
& Clock & 2 & $\checkmark$ & $\times$ \\
\bottomrule
\end{tabular}
\end{table}
\begin{table}[tbp]
\centering
\small
\caption{Policy outputs.}
\label{tab:policy_outputs}
\setlength{\tabcolsep}{2.6pt}\renewcommand{\arraystretch}{1.08}
\begin{tabular}{@{}>{\raggedright\arraybackslash}p{.18\linewidth}>{\raggedright\arraybackslash}p{.23\linewidth}rcccc@{}}
\toprule
\textbf{Output type} & \textbf{Joint / Signal} &
\shortstack[r]{\textbf{Res Limit}\\\textbf{[N$\cdot$m]}} &
\textbf{$K_p$} & \textbf{$K_d$} &
\textbf{RL} & \textbf{IL} \\
\midrule
\multirow{5}{*}{Leg joints}
& Hip yaw   & 18  & 77  & 23  & $\checkmark$ & $\checkmark$ \\
& Hip roll  & 160 & 255 & 77  & $\checkmark$ & $\checkmark$ \\
& Hip pitch & 27  & 110 & 33  & $\checkmark$ & $\checkmark$ \\
& Knee pitch& 220 & 420 & 105 & $\checkmark$ & $\checkmark$ \\
& Ankle roll& 28  & 120 & 20  & $\checkmark$ & $\checkmark$ \\[2pt]
Auxiliary
& Clock signal (gait phase) & --- & --- & --- & $\checkmark$ & $\times$ \\
\bottomrule
\end{tabular}
\end{table}

\begin{table}[tbp]
\centering
\small
\caption{Terrain anchor set used for pseudo-expert templates.}
\label{tab:terrain_anchors}
\setlength{\tabcolsep}{4pt}\renewcommand{\arraystretch}{1.08}
\begin{tabular}{@{}lcc>{\raggedright\arraybackslash}p{.44\linewidth}@{}}
\toprule
\textbf{} & \textbf{$\sigma$} & \textbf{Amp. (m)} & \textbf{Physical meaning} \\
\midrule
A1 & 0.65 & 0.06 & Mild roughness; low undulation \\
A2 & 0.65 & 0.12 & Mild roughness; larger undulation \\
A3 & 0.75 & 0.06 & Moderate roughness; low undulation \\
A4 & 0.75 & 0.12 & Moderate roughness; medium undulation \\
A5 & 0.85 & 0.08 & Smoother scale; small undulation \\
A6 & 0.85 & 0.15 & Smoother scale; large undulation \\
A7 & 0.75 & 0.10 & Central anchor (default) \\
\bottomrule
\end{tabular}
\end{table}

\noindent\textbf{Dataset disclosure.}
Robot reference trajectories are collected by pre-training a TRPO policy on multi-terrain environments (walk task, target speed 1.25~m/s, episode horizon 1000), with a maximum budget of $4\times10^7$ environment steps per seed (the realized steps may vary slightly with training progress). These trajectories are used only to construct the robot action library, rather than serving as final supervision.
Real demonstrations are loaded at 500~Hz and interpolated to the 40~Hz control frequency, yielding approximately $8.98\times10^4$ valid state transitions for the expert buffer.
Pseudo-expert trajectories are generated by the proposed two-tower Transformer + FGW projection: task-relevant temporal embeddings are learned for human and robot, FGW is solved in the shared embedding space, and barycentric projection maps the soft correspondence into executable robot actions.
To cover the continuous terrain parameterization used during training $(\sigma,\text{amplitude})$, we define seven representative terrain anchors (Table~\ref{tab:terrain_anchors}) and generate one pseudo-expert template per anchor. Each anchor corresponds to a specific terrain condition used to construct the pseudo-expert action library: A1--A2 represent mildly rough terrains (e.g., \textit{Level ground}); A3--A4 represent moderately rough terrains (e.g., \textit{Undulating turf}); A5--A6 capture larger-scale variations (e.g., \textit{Cobblestone} and \textit{Inclined ramp}); and A7 serves as the central reference template in the terrain parameter space, used as the default when interpolating intermediate conditions.
During training, these seven templates form the basis of the robot action library. When a terrain with parameters $(\sigma,\text{amplitude})$ is sampled, the corresponding pseudo-expert trajectory is obtained via barycentric interpolation among the relevant anchor templates rather than duplicating training data. Each template contains approximately $4.49\times10^5$ raw frames (corresponding to $\sim8.98\times10^4$ effective transitions after environment conversion).

\begin{figure}[tbp]
\centering
\includegraphics[width=1\linewidth]{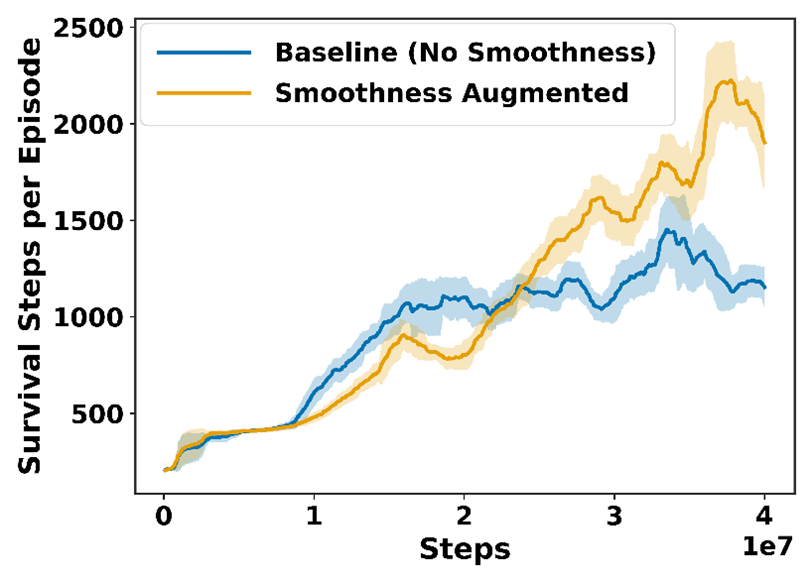}
\caption{Effect of smoothness reward on survival steps per episode.}
\label{fig:action_smoothing}
\end{figure}

\subsection{Multi-Terrain Applicability: Stability and Deployability}
To assess the generality of the pseudo-expert generation framework, we project human flat-ground demonstrations onto a half-body robot across four terrains---Level Ground, Undulating Turf, Cobblestone, and Inclined Ramp---and deploy a unified gait controller in each case. As shown in Fig.~\ref{fig:locomotion}, the pelvis-vertical trajectories exhibit terrain-specific patterns yet maintain rhythmic stability: periodic on level ground, slowly varying on turf, irregular but bounded on cobblestone, and smoothly graded on ramps. These results indicate that the generated pseudo-expert trajectories enable stable policy execution across all terrains, demonstrating cross-terrain applicability and effectiveness.
\subsection{Action-Smoothing Reward for RL Collection: Effects on Adaptation}
To test whether the new smoothing reward helps, we keep the training process and the network the same and change only whether we add action and torque smoothing. We measure performance by the number of survival steps in each episode until failure. Fig.~\ref{fig:action_smoothing} shows that the policy with smoothing gets higher survival steps across training, and the gain is larger in the middle and late stages.
\begin{figure*}[t]
\centering
\begin{minipage}[t]{.485\linewidth}
\centering
\includegraphics[width=\linewidth]{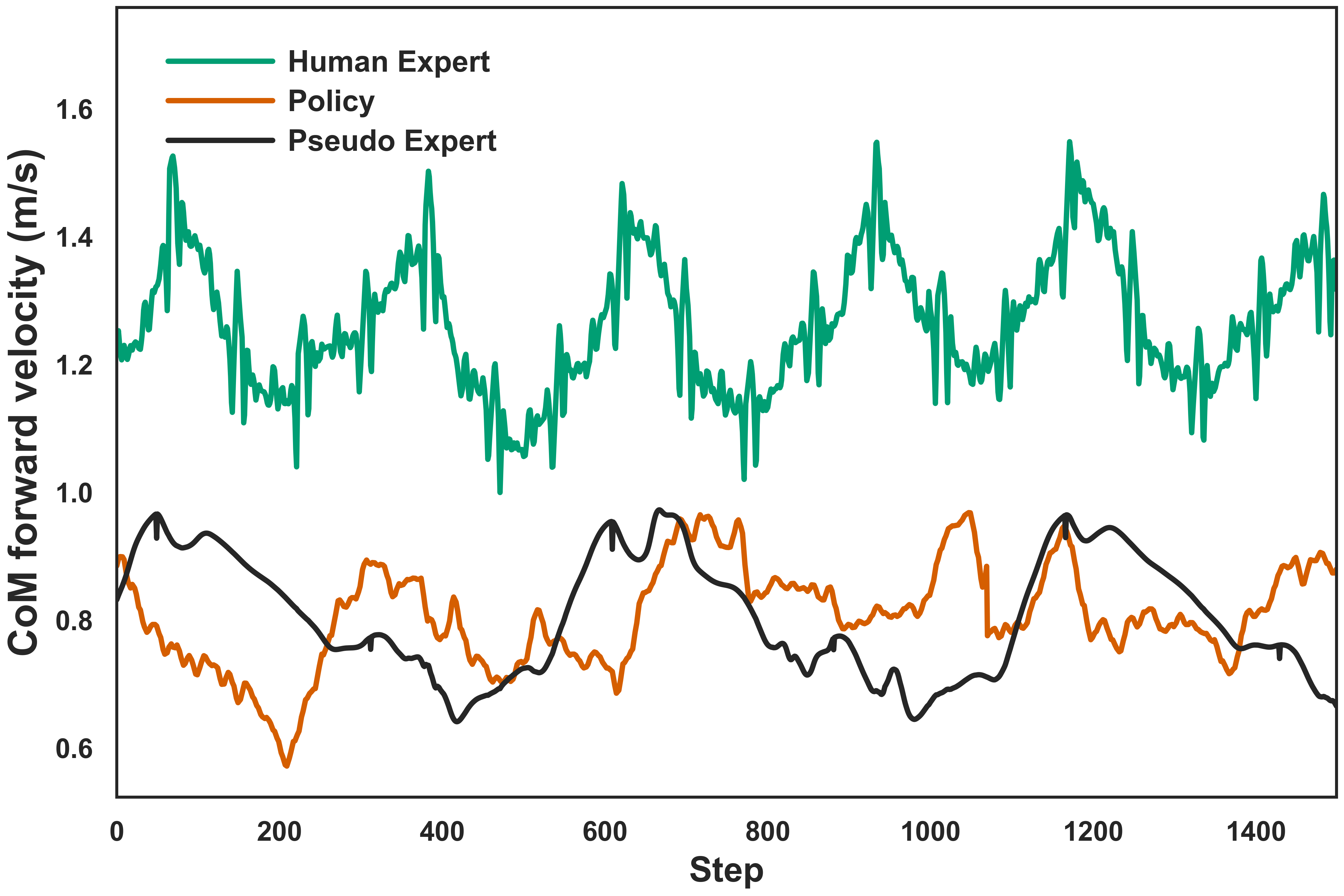}
\par\small (a) CoM forward velocity
\end{minipage}
\hfill
\begin{minipage}[t]{.485\linewidth}
\centering
\includegraphics[width=\linewidth]{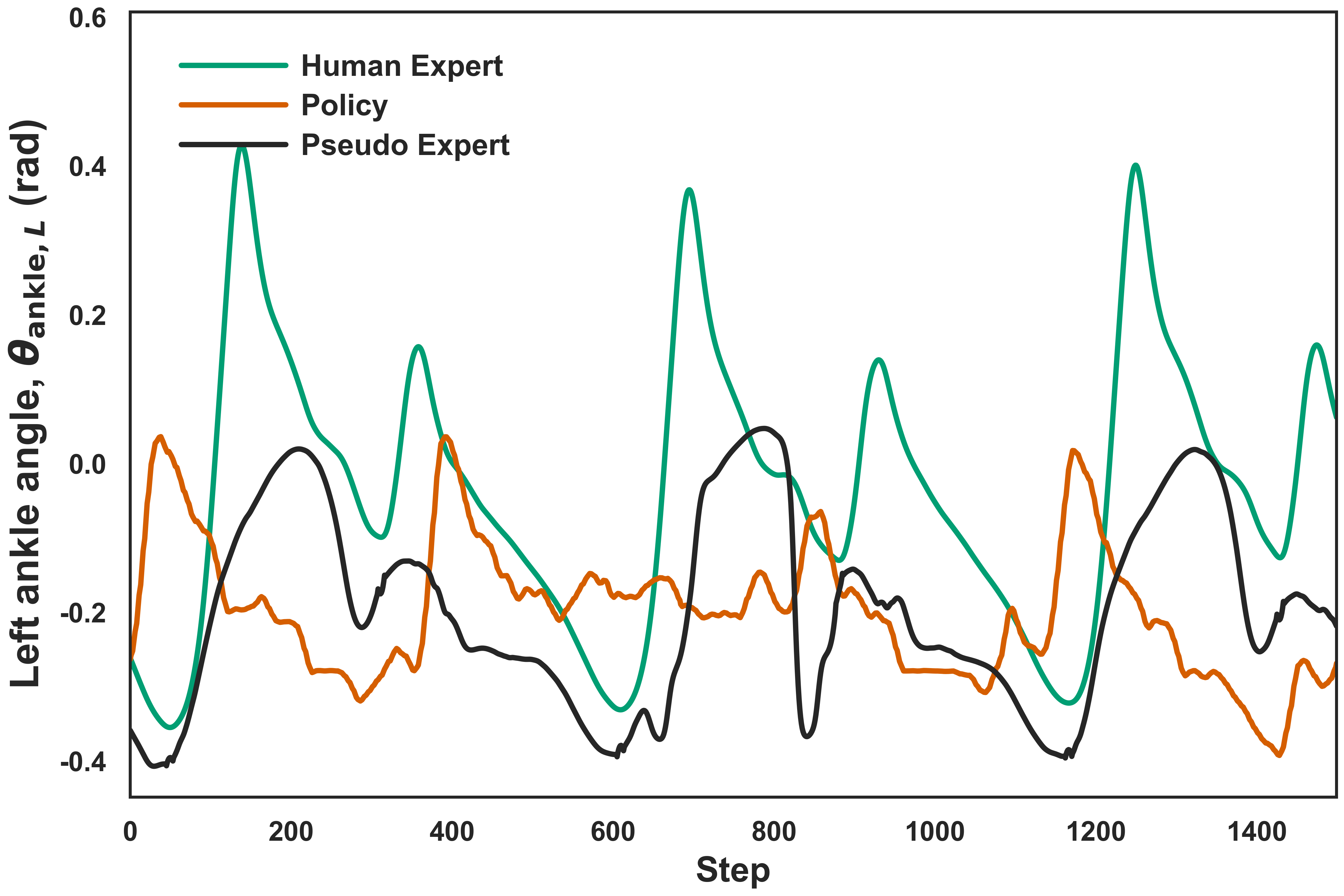}
\par\small (b) Left ankle joint angle
\end{minipage}
\caption{Comprehensive comparative analysis of CoM velocity and ankle joint-angle trajectories under pseudo-expert and human-expert supervision.}
\label{fig:fig4_ab}
\end{figure*}
\subsection{Pseudo-Expert Reproduction: Tracking Projected Trajectories}
\begin{figure}[tbp]
\centering
\includegraphics[width=1\linewidth]{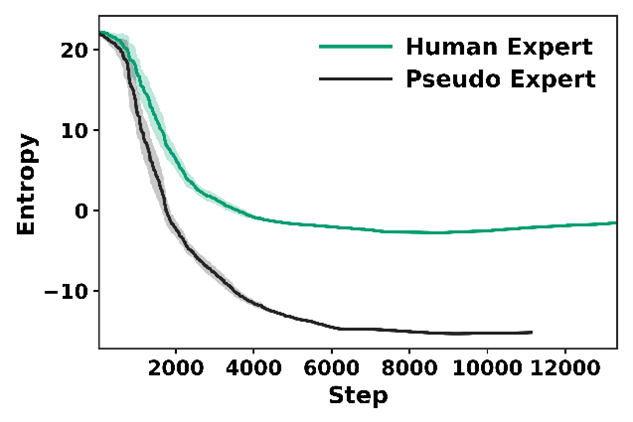}
\caption{Training dynamics: policy entropy under pseudo-expert vs.\ human supervision.}
\label{fig:fig5_entropy}
\end{figure}

To evaluate how well the policy reproduces expert motion, we compare Human Expert (green), Pseudo Expert (black), and Policy (orange) trajectories on level ground. As shown in Fig.~\ref{fig:fig4_ab}(a), using the Center of Mass (CoM) forward velocity, the Human Expert and the Policy exhibit bias in mean, amplitude, and phase, whereas the Pseudo Expert and the Policy match closely in mean, amplitude, and peak--trough timing. Fig.~\ref{fig:fig4_ab}(b) reports the left ankle angle $\theta_{\mathrm{ankle,L}}$, showing that the projection compresses the waveform adaptively and aligns its magnitude and phase with the Policy, reducing morphology- and dynamics-induced bias. The training dynamics (Fig.~\ref{fig:fig5_entropy}) agree with these observations: with pseudo-expert supervision, policy entropy drops rapidly and remains low, while direct human demonstrations lead to higher and fluctuating entropy. Overall, the Policy reproduces the pseudo-expert trajectories stably and remains executable.

\subsection{Relative Advantage of Pseudo-Expert}
\begin{figure*}[t]
\centering
\begin{minipage}[t]{.325\linewidth}
\centering
\includegraphics[width=\linewidth]{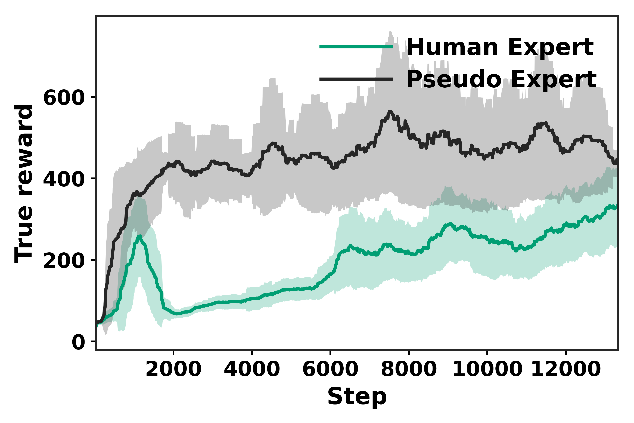}
\par\small (a) True reward vs. step
\end{minipage}
\hfill
\begin{minipage}[t]{.325\linewidth}
\centering
\includegraphics[width=\linewidth]{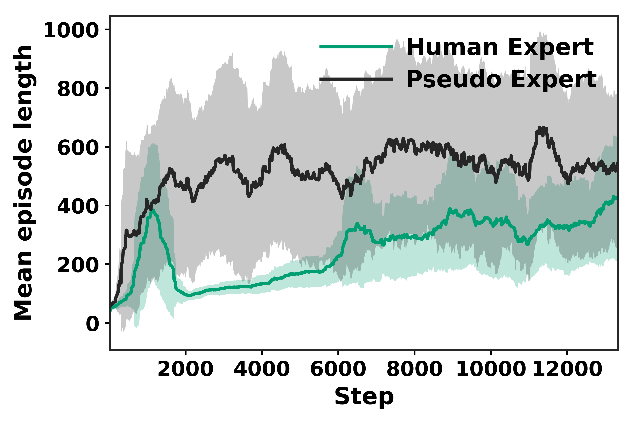}
\par\small (b) Mean episode length vs. step
\end{minipage}
\hfill
\begin{minipage}[t]{.325\linewidth}
\centering
\includegraphics[width=\linewidth]{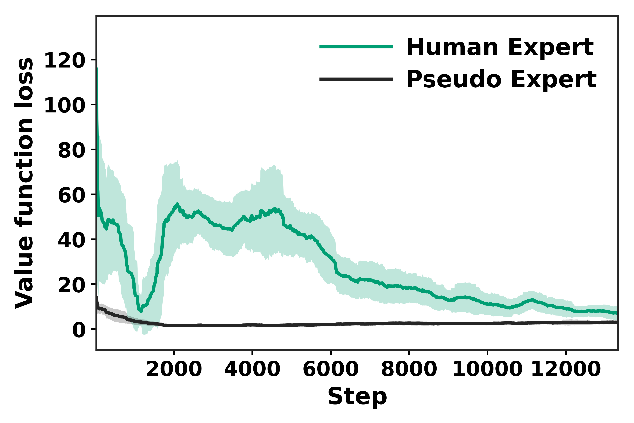}
\par\small (c) Value function loss vs. step
\end{minipage}
\caption{Learning curves on flat ground (Pseudo-Expert vs. Human Expert).}
\label{fig:fig6_curves}
\end{figure*}

We assess whether pseudo-expert trajectories outperform direct human demonstrations. We train policies on a flat-ground walking task under two supervision settings--- Pseudo Expert and Human Expert---using Variational Adversarial Imitation Learning (VAIL). We evaluate three aspects: true reward (Fig.~\ref{fig:fig6_curves}a), average episode length (Fig.~\ref{fig:fig6_curves}b), and value-function loss (Fig.~\ref{fig:fig6_curves}c).First, Fig.~\ref{fig:fig6_curves}a shows that the pseudo-expert setting enters the high-reward regime earlier and attains a higher overall level, which indicates better sample efficiency and a higher performance ceiling. Second, Fig.~\ref{fig:fig6_curves}b shows that the pseudo-expert setting increases episode length faster and maintains it at a higher level for longer, which implies fewer early terminations and more persistent gait stability. Finally, Fig.~\ref{fig:fig6_curves}c shows that the pseudo-expert setting yields a lower and smoother value-function loss, which suggests a more stable critic fit and a smaller estimation variance. Taken together, the closer state--action alignment between pseudo-expert data and the platform dynamics reduces learning noise and distribution mismatch, thereby enabling faster convergence and better task performance.

\subsection{Embedding-Space FGW versus Raw-Space FGW}

To isolate the effect of solving FGW in the learned embedding space, we compare two projection variants under identical Projection--Training settings: (i) \textit{Raw-space FGW}, which computes FGW directly in the original trajectory/state representation, and (ii) \textit{Embedding-space FGW} (ours), which computes FGW in the shared task-relevant embedding learned by the two-tower encoder. All other components (imitation objective, optimizer, and training protocol) are kept unchanged.

\begin{figure*}[t]
\centering
\begin{minipage}[t]{.485\linewidth}
\centering
\includegraphics[width=\linewidth]{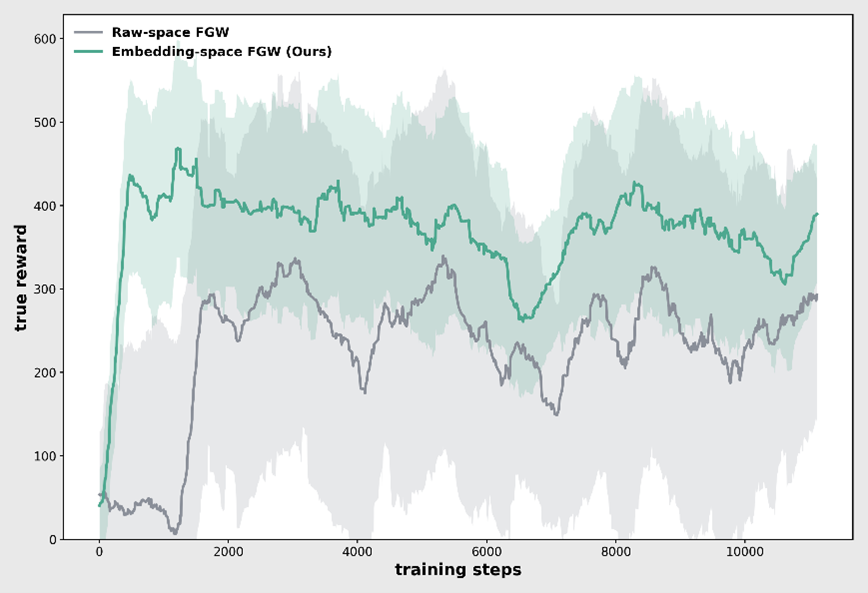}
\par\small (a) True reward vs.\ training steps.
\end{minipage}
\hfill
\begin{minipage}[t]{.485\linewidth}
\centering
\includegraphics[width=\linewidth]{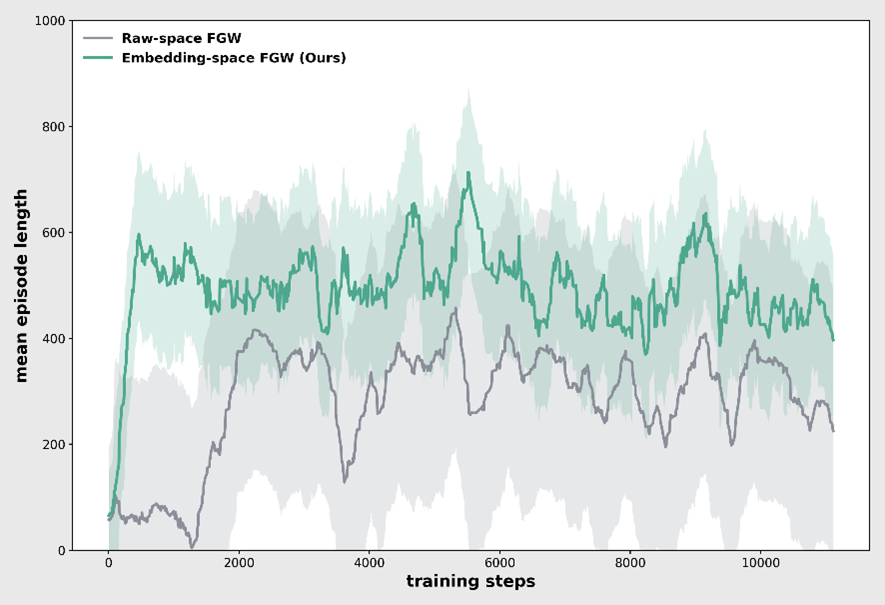}
\par\small (b) Mean episode length vs.\ training steps.
\end{minipage}
\caption{Embedding-space FGW vs.\ raw-space FGW under identical settings. Solving FGW in the shared embedding space reaches high-reward regimes earlier and maintains longer episode lengths throughout training.}
\label{fig:fgw_raw_vs_emb}
\end{figure*}

As shown in Fig.~\ref{fig:fgw_raw_vs_emb}, solving FGW in the learned embedding space consistently outperforms solving it in the raw trajectory/state space in both reward (a) and episode length (b). Embedding-space FGW enters the high-performance regime much earlier and maintains a higher operating band over the full training horizon, while Raw-space FGW improves later and remains at a lower level with more frequent regressions. The gap is particularly pronounced in the early training stage, where the embedding-space variant rapidly escapes the low-reward region, suggesting substantially better sample efficiency and a more informative initial alignment. Moreover, although both variants exhibit oscillations due to adversarial optimization and terrain variability, the embedding-space formulation shows stronger recovery after temporary drops, indicating improved training stability. This indicates that the shared task-relevant embedding suppresses embodiment-specific distortions in raw space and yields more reliable correspondences, producing pseudo-expert supervision that is more executable, less noisy, and better matched to downstream adversarial imitation learning. Consequently, the performance gain stems not merely from using FGW, but from solving FGW in an appropriately structured representation space.

\subsection{Dependency on Initial RL Data Quality}

The generation of pseudo-expert trajectories relies on an initial trajectory library collected by a reinforcement learning (RL) policy. To investigate how the quality of this initial policy affects downstream imitation performance, we conduct an ablation study by deliberately degrading the RL data collection process.

Specifically, we compare three RL training budgets used for trajectory collection: $1\times10^7$, $2\times10^7$, and $4\times10^7$ samples, where $4\times10^7$ corresponds to our default configuration. The resulting trajectory libraries are then used to construct pseudo-expert supervision under identical Projection--Training settings.

\begin{figure*}[t]
\centering
\begin{minipage}[t]{.485\linewidth}
\centering
\includegraphics[width=\linewidth]{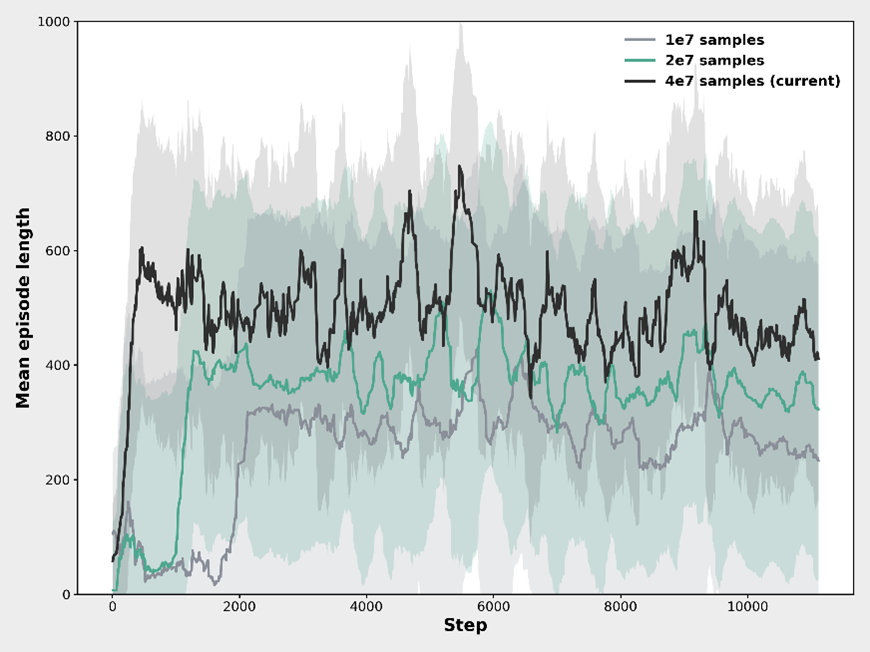}
\par\small (a) Mean episode length under different RL data budgets.
\end{minipage}
\hfill
\begin{minipage}[t]{.485\linewidth}
\centering
\includegraphics[width=\linewidth]{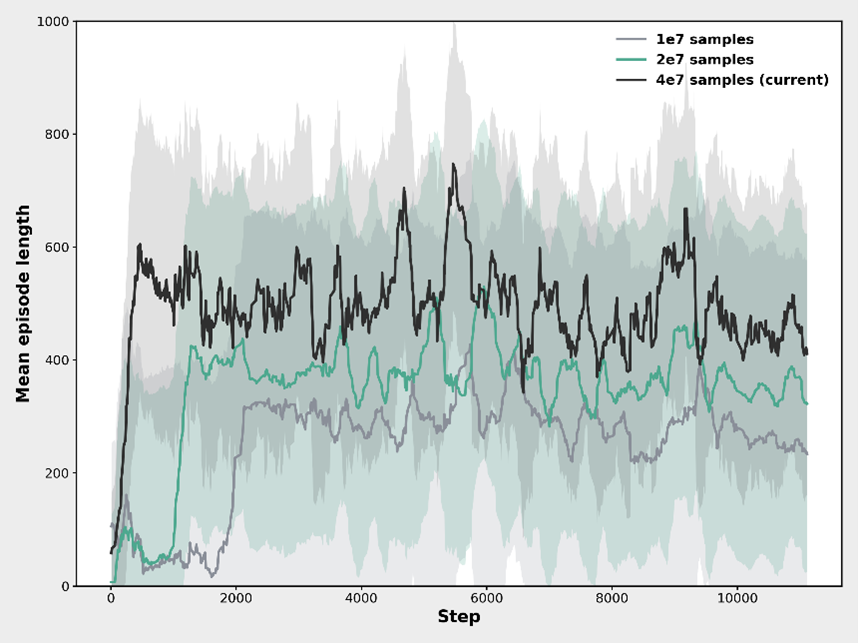}
\par\small (b) Training stability comparison under degraded RL initialization.
\end{minipage}
\caption{Effect of initial RL trajectory library quality on downstream imitation learning.}\label{fig:rl_data_quality}
\end{figure*}

As shown in the figures, improving the RL training budget leads to higher stability and longer mean episode lengths during imitation learning. Importantly, even with degraded RL initialization ($1\times10^7$ samples), the pseudo-expert supervision remains stable and does not collapse, indicating that the projection mechanism is robust to moderate degradation in the initial trajectory library.

\subsection{Adaptive Curriculum--- Effects on Generalization and Unseen-Terrain Robustness}
\begin{figure*}[t]
\centering
\includegraphics[width=.94\linewidth]{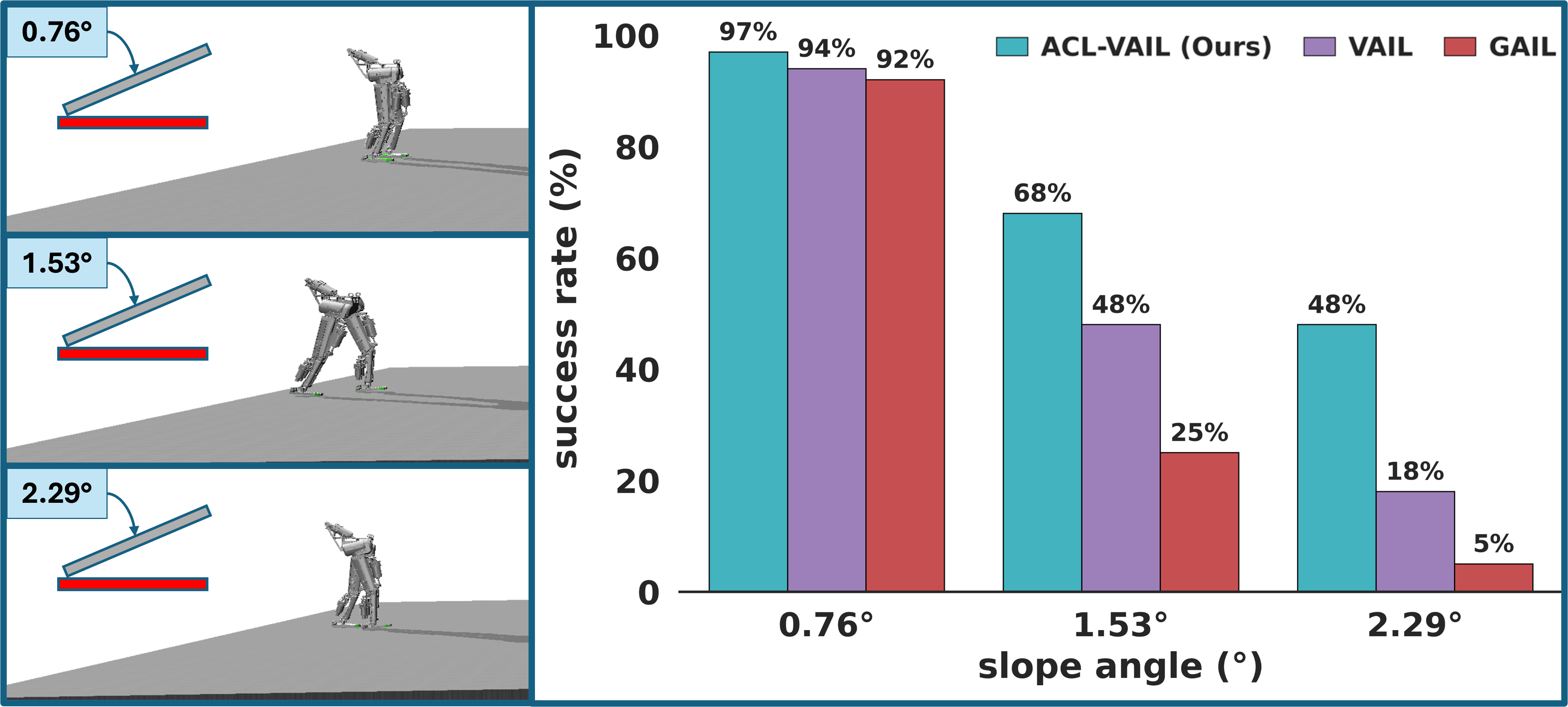}
\caption{Robustness under slope extrapolation (ID~0.76$^\circ$, OOD~1.53$^\circ$/2.29$^\circ$): ACL-VAIL vs.\ VAIL vs.\ GAIL.}
\label{fig:slope_robustness}
\end{figure*}

To test whether ACL and prioritized resampling improve robustness, we compare three IL variants under a unified pipeline: GAIL (Generative Adversarial Imitation Learning), VAIL, and ACL-VAIL (VAIL + ACL + prioritized resampling).%
In a symmetric-slope environment (Fig.~\ref{fig:slope_robustness}) with angles $0.76^{\circ}$ (ID) and $1.53^{\circ}$, $2.29^{\circ}$ (OOD), all methods use pseudo-experts and the same discriminator, with demonstrations only at $0.76^{\circ}$.%
The metric is success rate (non-fall rollouts / 100).%
Results (ACL-VAIL / VAIL / GAIL): $0.76^{\circ}$---97/94/92\%, $1.53^{\circ}$---68/48/25\%, $2.29^{\circ}$---48/18/5\%.%
ACL-VAIL slows degradation as slope increases and shows stronger cross-distribution generalization and robustness.%

\subsubsection{Ablation: Terrain Randomization and Prioritized Resampling}

\begin{figure}[tbp]
\centering
\includegraphics[width=1\linewidth]{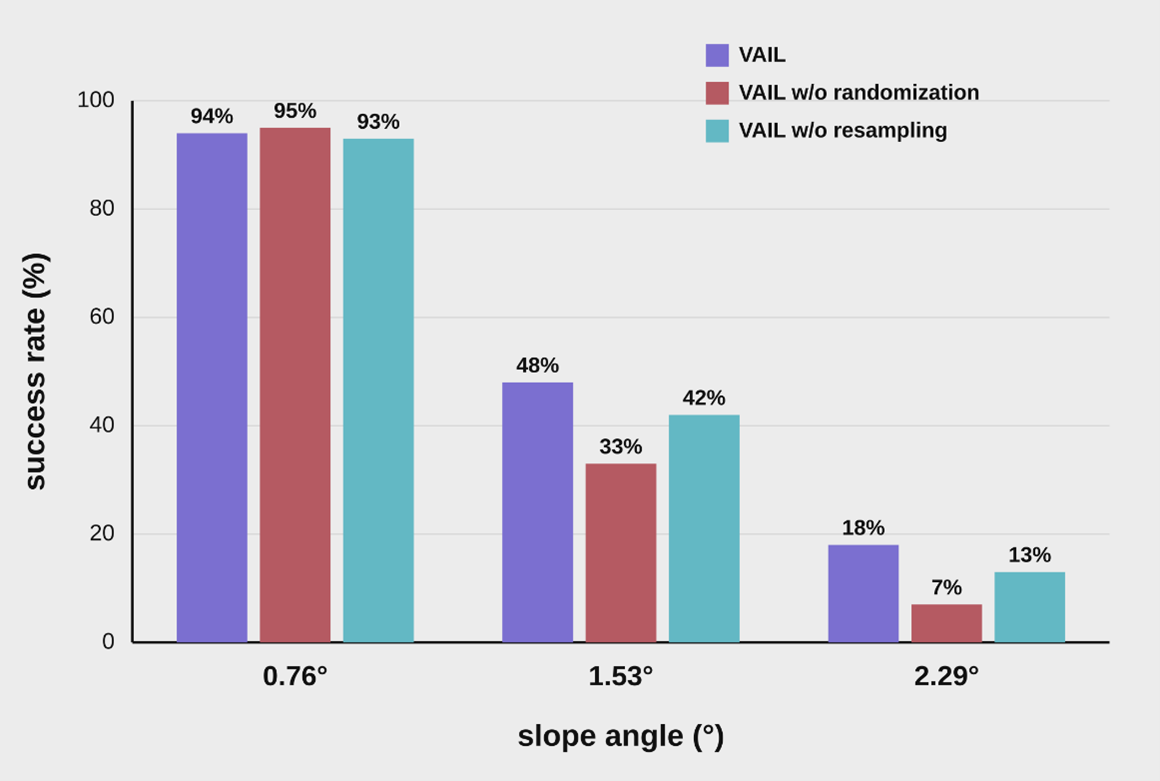}
\caption{Ablation of curriculum terrain randomization and prioritized resampling under slope extrapolation.}
\label{fig:abl_rand_resamp}
\end{figure}

To disentangle the impact of curriculum terrain randomization and prioritized resampling, we run three variants under the same pseudo-expert supervision, identical training budgets, and identical random seeds: (i) VAIL (full), (ii) VAIL w/o randomization (terrain randomization disabled), and (iii) VAIL w/o resampling (success-band prioritized resampling disabled).
As shown in Fig.~\ref{fig:abl_rand_resamp}, the three variants behave similarly under the ID condition (0.76$^\circ$ slope), where all of them achieve success rates above 90\%. This indicates that, within the training distribution, the pseudo-expert supervision and the base VAIL objective are already sufficient to support stable locomotion, and neither terrain randomization nor prioritized resampling is critical for solving the nominal task itself. In fact, the small advantage of the no-randomization variant in-distribution suggests that removing environment diversity may slightly simplify the optimization problem when generalization is not required.

The difference becomes substantial once the slope moves outside the training distribution. At 1.53$^\circ$, disabling terrain randomization reduces the success rate from 48\% to 33\%, while disabling prioritized resampling yields an intermediate value of 42\%. At 2.29$^\circ$, the gap widens further: the full method reaches 18\%, compared with only 7\% without randomization and 13\% without resampling. These results show that terrain randomization is the primary contributor to cross-slope generalization, because it broadens the terrain distribution experienced during training and reduces over-specialization to narrow conditions. Prioritized resampling provides an additional but secondary benefit by reinforcing successful behaviors near the boundary of the current capability range, which helps slow down performance degradation as difficulty increases.

Taken together, the two mechanisms are complementary rather than redundant. Terrain randomization expands the support of the training distribution, while prioritized resampling sharpens learning within challenging but learnable regions of that distribution. Their combination therefore produces the slowest degradation trend under OOD slopes and yields the most robust extrapolation behavior.

\subsection{Cross-Morphology Generalization}

To further verify that the proposed Projection--Training framework is not restricted to a half-body humanoid morphology, we directly transfer the entire pipeline to a full-body humanoid platform (Unitree H1) without modifying the Transformer encoder, FGW projection mechanism, or imitation-learning architecture.

Under identical Projection--Training settings, pseudo-expert trajectories are generated via the same FGW projection and used to supervise imitation learning on the full-body robot. No morphology-specific joint remapping or architectural redesign is introduced, ensuring a fair cross-morphology validation.

\begin{figure}[tbp]
\centering
\includegraphics[width=1\linewidth]{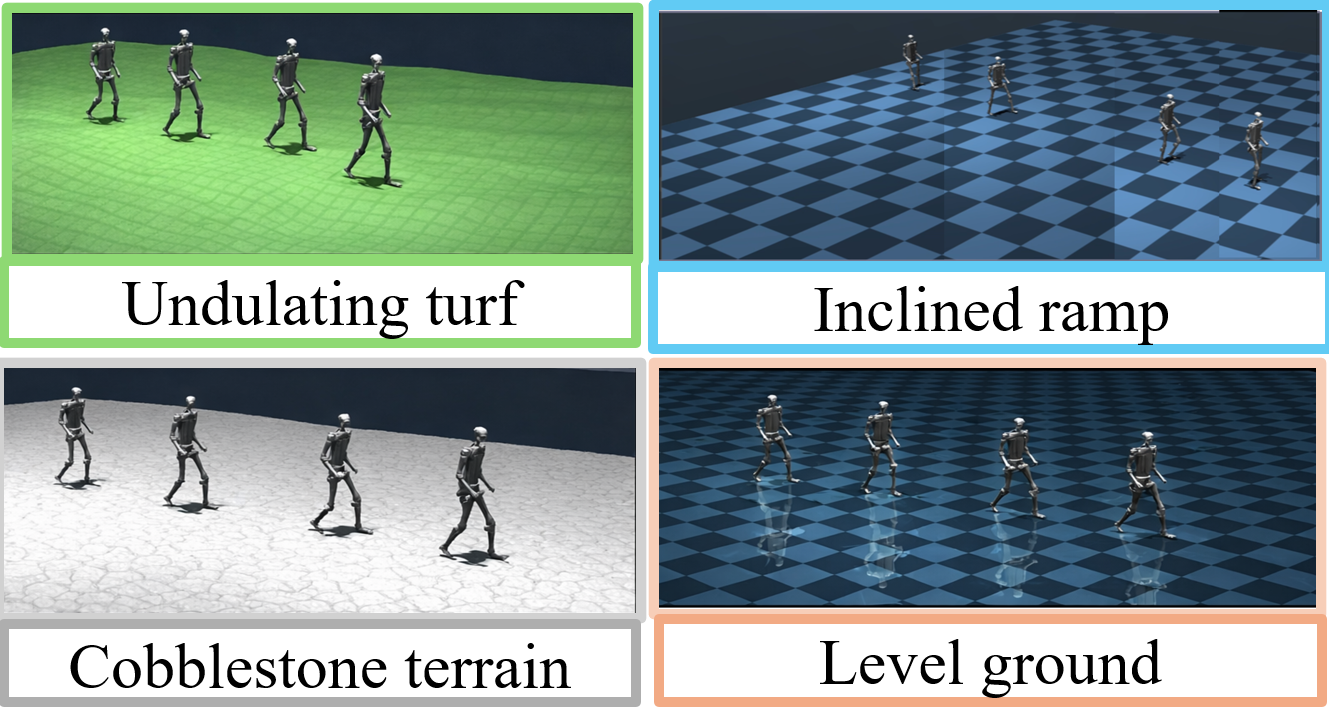}
\caption{Cross-morphology transfer results on a full-body humanoid (Unitree H1). The policy trained with the identical Projection--Training pipeline achieves stable locomotion without structural modification.}
\label{fig:unitree}
\end{figure}

These results demonstrate that the shared Transformer encoder and FGW-based projection operate in a morphology-agnostic embedding space and are not tied to a specific kinematic structure.

\section{Conclusion}
This paper presented TRAIL, a projection-before-training framework for cross-embodiment imitation learning under terrain variation. Instead of learning policies directly from embodiment-mismatched human demonstrations, TRAIL first learns task-relevant temporal embeddings for human and robot motion, solves Fused Gromov--Wasserstein (FGW) correspondence in the shared embedding space, and converts the resulting soft alignment into executable pseudo-expert supervision via barycentric projection. This design decouples correspondence estimation from policy optimization, preserves the imitation-learning paradigm, and reduces reliance on embodiment-matched demonstrations.

Extensive simulation results support both the effectiveness of this design and the role of its individual components. On the half-body humanoid, pseudo-expert supervision yields faster convergence, longer episode lengths, lower critic loss, lower policy entropy, and closer trajectory tracking than direct human supervision. Solving FGW in the learned embedding space is substantially more effective than solving it in the raw state space, validating the proposed alignment formulation. Experiments with degraded RL trajectory libraries further show that the method remains stable under moderate reductions in initial data quality, although higher-quality libraries improve the performance ceiling. Under slope extrapolation, terrain randomization contributes more to out-of-distribution robustness, while prioritized resampling further mitigates performance degradation. Cross-morphology transfer to a full-body humanoid (Unitree H1) under the same Projection--Training pipeline indicates that the learned alignment is not tied to a single half-body morphology.

Overall, TRAIL provides an effective way to transform mismatched demonstrations into executable supervision for humanoid locomotion across terrain and morphology changes. Future work will extend the framework to physical robot deployment, broader robot embodiments, and more diverse real-world demonstration sources.

\section*{Acknowledgements}
The authors thank the editors and reviewers for their comments, which led to improvement of this paper. This work is supported by the Opening Foundation of Key Laboratory of Intelligent Robot(No.HBIR 202507)

\balance
\end{document}